%% file: main.tex
\documentclass{article}
\usepackage{booktabs}
\usepackage[colorlinks=true,citecolor=blue,anchorcolor=purple]{hyperref}
\usepackage{url}
\usepackage{multirow}
\usepackage{balance}
\usepackage{spconf,amsmath,graphicx, booktabs}

\renewenvironment{thebibliography}[1]{
  \begin{oldthebibliography}{#1}
    \setlength{\itemsep}{0.3ex}
    \setlength{\parskip}{0.3ex}
}
{
  \end{oldthebibliography}
}
 \usepackage{relsize}
 \usepackage{amssymb}

\title{Activation Compression of Graph Neural Networks using Block-wise Quantization with Improved Variance Minimization}

\name{Sebastian Eliassen$^{\dagger}$ \qquad Raghavendra Selvan$^{\dagger}$\sthanks{The authors acknowledge funding received under European Union’s Horizon Europe Research and Innovation programme under grant agreements No. 101070284 and No. 101070408.}}

\address{$^{\dagger}$  Department of Computer Science, University of Copenhagen}
\input{Macros}
\begin{document}
%
\maketitle
%


\begin{abstract}
Efficient training of large-scale graph neural networks (GNNs) has been studied with a specific focus on reducing their memory consumption. Work by Liu et al. (2022) proposed extreme activation compression (EXACT) which demonstrated drastic reduction in memory consumption by performing quantization of the intermediate activation maps down to using {\tt INT2} precision. They showed little to no reduction in performance while achieving large reductions in GPU memory consumption. In this work, we present an improvement to the EXACT strategy by using block-wise quantization of the intermediate activations. We experimentally analyze different block sizes and show further reduction in memory consumption ($> 15\%$), and runtime speedup per epoch ( $\approx 5\%$) even when performing extreme extents of quantization with similar performance trade-offs as with the original EXACT. Further, we present a correction to the assumptions on the distribution of intermediate activation maps in EXACT (assumed to be uniform) and show improved variance estimations of the quantization and dequantization steps. \footnote{Source code is available at the official paper repository \url{https://github.com/saintslab/i-Exact}.}
\end{abstract}
\begin{keywords}
graph neural networks, quantization, activation compression, efficient machine learning, deep learning
\end{keywords}

\section{Introduction}
Graph neural networks (GNNs) are a class of deep learning (DL) models most useful when dealing with graph structured data~\cite{scarselli2008graph,gcns}. They have shown widespread relevance in a range of diverse applications~\cite{do2019matrix,zhao2021distributed,tzirakis2021multi,cervino2023training}. GNNs are known to scale poorly with the number of nodes in the graph data primarily due to the memory requirements for storing the adjacency matrices and intermediate activation maps~\cite{duan2022comprehensive}. The increase in memory consumption necessitates use of more computational resources. This is, unfortunately, in line with the growing resource consumption of recent classes of deep learning methods~\cite{anthony2020carbontracker,sevilla2022compute}. A common approach to reducing resource consumption of DL methods is to explore different efficiency strategies~\cite{bartoldson2023compute, wright2023efficiency} such as training neural networks with quantized weights~\cite{hubara2016binarized} or quantized activation maps~\cite{chen2021actnn}. 

The main focus of efficiency improvements in GNNs has been either by operating on subgraphs to use smaller adjacency matrices~\cite{hamilton2017inductive} or to store compressed node embeddings or activation maps for computing gradients~\cite{exact}. In this work, we are interested in the latter, specifically following the method introduced in~\cite{exact} that proposed extreme activation compression (EXACT) using a combination of stochastic rounding-based quantization and random projections. 

In this work we make two contributions, starting from EXACT, that further improve the memory consumption and yield training runtime speedup. Firstly, we introduce block-wise quantization~\cite{blockwise} of the activation maps which quantizes large groups of tensors instead of individual tensors with support down to {\tt INT2} precision. Secondly, the quantization variance estimation in EXACT is performed using assumption that the activation maps uniformly distributed. We show that the activation maps do not follow a uniform distribution but instead follow a type of clipped normal distribution with empirical evidence. Using this insight, we present an improvement to the variance minimization strategy when performing the quantization of activation maps. Experimental evaluation on multiple graph datasets shows a consistent reduction in memory consumption and speedup in training runtime compared to EXACT. 

\vspace{-0.25cm}
\section{Notations and Background}
\label{sec:back}
\vspace{-0.25cm}
We describe a graph with $N$ nodes as $\Gcal = (\Xbf, \Abf)$, with dense node feature matrix $\Xbf \in \Rm^{N \times F}$ containing $F$-dimensional features for each of the $N$ nodes, and sparse adjacency matrix $\Abf \in \{0, 1\}^{N \times N}$ with the relations between each of the nodes. Specifically $\Abf_{i, j} = 1$ if there is an edge between node $i$ and $j$ and $\Abf_{i, j} = 0$ otherwise.

The GNN from~\cite{gcns} with $L$ layers can be compactly written as the recursion:
\begin{equation}
    \Hbf^{(\ell+1)} = \mathlarger{\sigma}\left(\hat\Abf\Hbf^{(\ell)}\mathbf{\Theta}^{{(\ell)}}\right) 
    \label{eq:quant}
\end{equation}
where the symmetric normalized adjacency matrix is $\hat\Abf = \tilde\Dbf^{-\frac{1}{2}}\Abf\tilde\Dbf^{-\frac{1}{2}}$ with $\tilde{\Dbf}$ as the degree matrix of $ \Abf + \Ibf$, $\Hbf^{(0)} := \Xbf$, the trainable parameters at layer-$\ell$ are $\mathbf{\Theta}^{(\ell)}$ and a suitable non-linearity $\mathlarger{\sigma}(\cdot)$. 

\begin{figure}[t]
    \centering
    \includegraphics[width=0.239\textwidth]{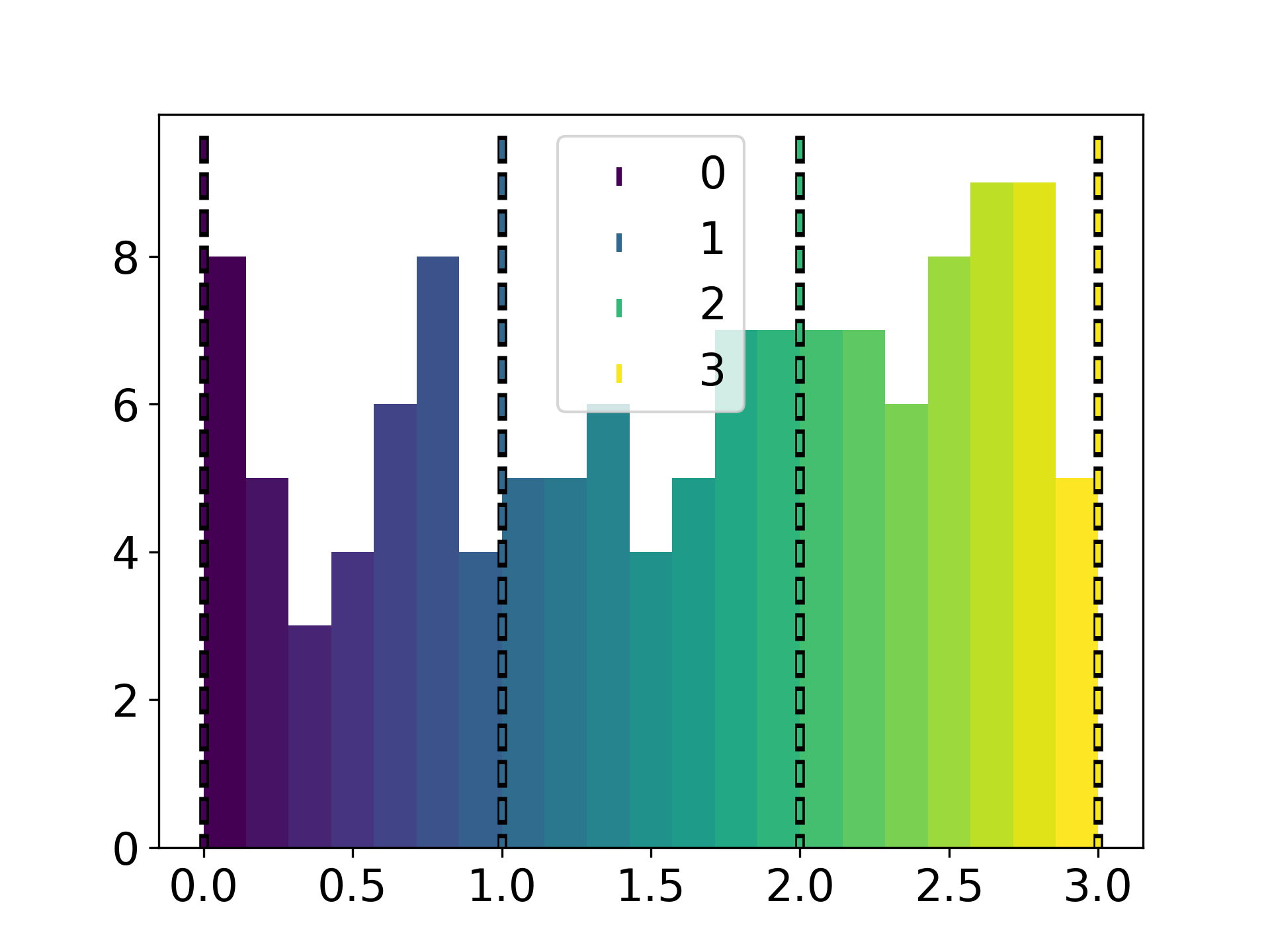}
    \includegraphics[width=0.239\textwidth]{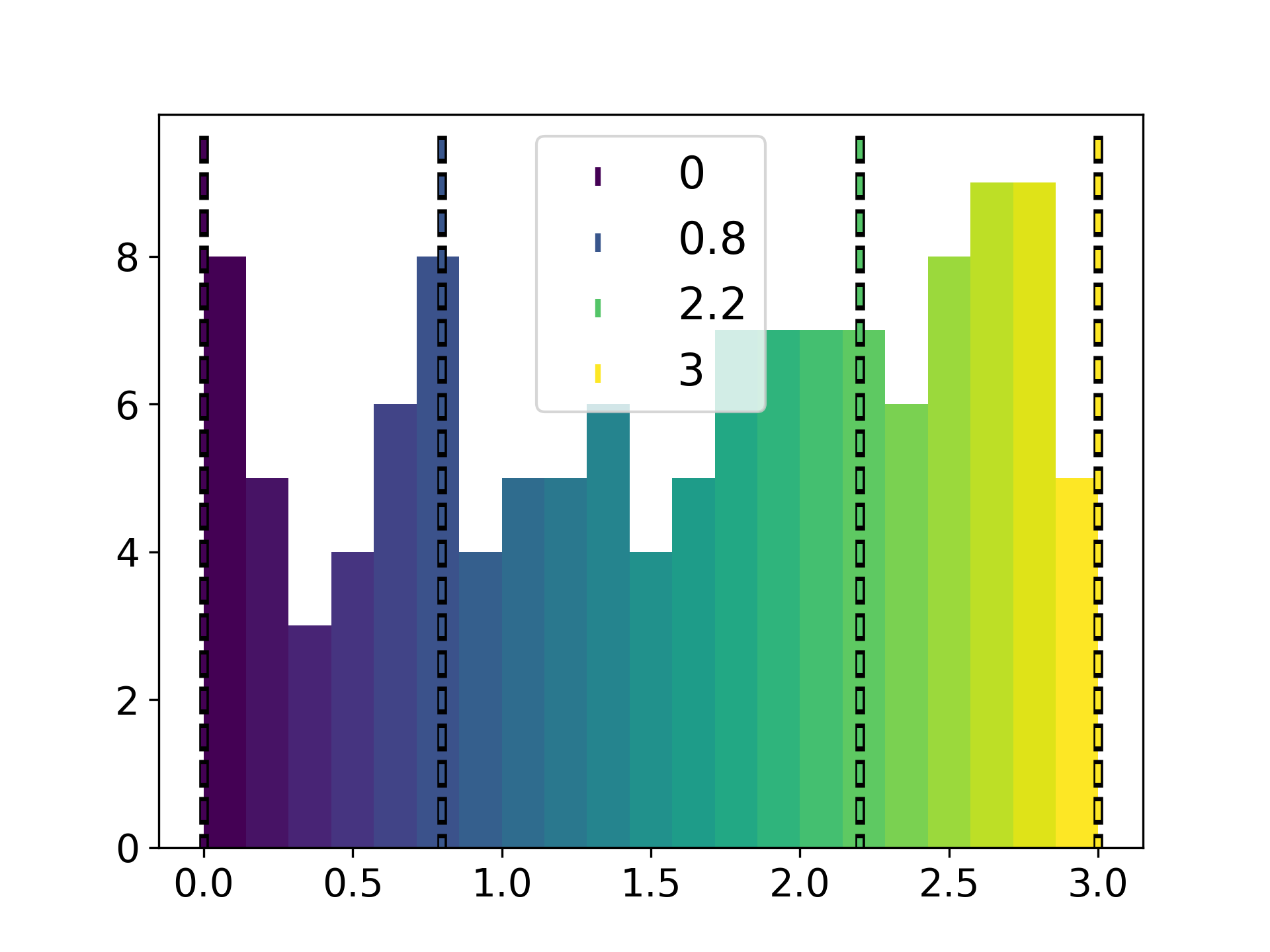}
    \vspace{-0.55cm}
    \caption{{\em Demonstration of stochastic rounding for $b=2$ i.e., $2^b=4$ quantization bins for 128 points uniformly sampled datapoints. Here the sampled points can be quantized to any of the four levels. The closer the color of the sample is to the color of the vertical bar, the larger the probability that it quantizes to the said vertical bar.  Quantization bins when using uniform bin widths (left) and when using non-linear bin widths when performing variance optimization (right) introduced in Sec~\ref{sec:variance} is visualized.}.}
    \label{fig:quant_demo}
    \vspace{-0.5cm}
\end{figure}
Since the activation maps, specifically the intermediate results $\left(\Hbf^{(\ell)}\mathbf{\Theta}^{{(\ell)}}\right)$ and the node embedding matrix $\Hbf^{(\ell)}$, are the biggest users of memory, EXACT~\cite{exact} focused on reducing the size of the activation maps from {\tt FLOAT32} to lower precision using two methods: \\
{\bf Stochastic Rounding}: For a given node $i$ its embedding vector $\hbf_{i}^{(\ell)}$ is quantized and stored using $b$-bit integers as:
\begin{equation}
    \hbf_{i_\text{\tt INT}}^{(\ell)} = \operatorname{Quant}\left(\hbf_{i}^{(\ell)}\right) = \left\lfloor \frac{\hbf_{i}^{(\ell)} - Z_i^{(\ell)}}{r_i^{(\ell)}}B \right\rceil = \left\lfloor \bar{\hbf} \right\rceil
\end{equation}
where $B=2^b-1$, $Z_i^{(\ell)} = \min(\hbf_{i}^{(\ell)})$ is the zero-point, $r_i^{(\ell)} = \max(\hbf_{i}^{(\ell)}) - \min(\hbf_{i}^{(\ell)})$ is the range for $\hbf_{i}^{(\ell)}$, $\bar{\hbf}$ is the normalized activation map,  and $\sr{\cdot}$ is the stochastic rounding (SR) operation~\cite{courbariaux2015binaryconnect}. SR rounds a number to its nearest integer with a probability inversely proportional to the distance from the quantization boundaries.\footnote{For any scalar activation map, $h$, SR is given by: 
\begin{equation*}
    \left\lfloor h \right\rceil = \begin{cases}
        \lfloor h \rfloor + \delta, \text{with probability } (h-\lfloor h \rfloor)/\delta\\
        \lfloor h \rfloor, \text{with probability } 1-(h-\lfloor h \rfloor)/\delta\\
    \end{cases}
\end{equation*}
where $\delta$ is the uniform bin width, $\lfloor \cdot \rfloor$ is the floor operator.} It can be showed that SR is an unbiased operator as the rounding probabilities are dependent on distance of $\left\lfloor \bar{\hbf} \right\rceil$ to the nearest integers~\cite{improved_rounding}.\footnote{
Consider, the expectation according to the definition of SR:
\begin{align*}
    \Em[\left\lfloor h \right\rceil] &= \left(\lfloor h \rfloor + \delta\right) \cdot \left(h-{\lfloor h \rfloor}\right)/{\delta} + \lfloor h \rfloor \cdot \left(1-\left(\left(h-{\lfloor h \rfloor}\right)/{\delta}\right)\right) \nonumber \\ 
    &= h-{\lfloor h \rfloor} + \lfloor h \rfloor = h.
\end{align*}
This proves that SR with uniform bin widths is unbiased.} Figure \ref{fig:quant_demo}-A) illustrates SR with uniform bin widths.

The inverse process of dequantization is defined as:
\begin{equation}
\hat{\hbf}_{i}^{(\ell)} = \operatorname{Dequant}\left(\hbf_{i_\text{\tt INT}}^{(\ell)}\right ) = r_i^{(\ell)} \hbf_{i_\text{\tt INT}}^{(\ell)} / B + Z_i^{(\ell)}    
\end{equation}
which linearly transforms the quantized values from $[0, B]$ back to their original ranges. Note that we still have some information-loss, since $\hbf_{i\text{\tt INT}}^{(\ell)}$ is only an estimate of $\hbf_{i}^{(\ell)}$.\footnote{Note that quantization followed by dequantization is unbiased due to stochastic rounding, i.e.,
$\Em[\hat{\hbf}_{i}^{(\ell)}] = \Em [\operatorname{Dequant}(\operatorname{Quant}(\hbf_{i}^{(\ell)}))] = {\hbf}_{i}^{(\ell)}.$}
\\
{\bf Random Projection}: Another way of minimizing memory footprint of activation maps is to perform dimensionality reduction on them. This is done via random projection in EXACT as: 
\begin{equation}
\hbf_{i_\text{\tt proj}}^{(\ell)} = \operatorname{RP}({\hbf}_{i}^{(\ell)}) = {\hbf}_{i}^{(\ell)}\Rbf
\end{equation}
where $\mathbf{R} \in \mathbb{R}^{D \times R}$ with $R < D$ is the normalized Rademacher random  matrix~\cite{rp_linden} that satisfies $\mathbb{E}[\mathbf{RR}^\top] = \mathbf{I}$. 

The random projected node embeddings are inversely transformed by 
\begin{equation}
\hat{\hbf}_{i}^{(\ell)} = \operatorname{IRP}\left(\hbf_{i_\text{\tt proj}}^{(\ell)}\right) = \hbf_{i_\text{\tt proj}}^{(\ell)}\Rbf^T.
\end{equation}
The matrix containing all projected and recovered activation maps are defined as $\mathbf{H}_\text{proj}^{(\ell)}$ and $\hat{\mathbf{H}}^{(\ell)}$, respectively.\footnote{Also note that the RP and IRP operations are also unbiased. i.e., $\Em[\hat{\mathbf{H}}^{(\ell)}] = \Em[\text{IRP}(\text{RP})(\mathbf{H}^{(\ell)})] = \mathbf{H}^{(\ell)}$.}

EXACT method combines random projection and quantization to obtain compounding reductions in memory consumption. Specifically, node embeddings are compressed as $\tilde{\hbf}_i^{(\ell)} = \operatorname{Quant}\left(\operatorname{RP}\left(\hbf_i^{(\ell)}\right)\right)$ are stored in memory during the forward pass, and during the backward pass the they are recovered as $\hat{\hbf}_i^{(\ell)} = \operatorname{IRP}\left(\operatorname{Dequant}\left(\tilde{\hbf}_i^{(\ell)}\right)\right)$. 



\section{Methods}
Quantizing activation maps of GNNs reduces the memory consumption when training GNNs but does introduce an additional overhead in the computation time due to the quantization/dequantization steps. We propose to perform large block-wise quantization~\cite{blockwise} in place of quantizing individual tensors in order to recover some of the slowdown and to further reduce the memory consumption. 

\subsection{Block-wise Quantization of Activation maps}

The quantization in Eq.~\eqref{eq:quant} is performed over each node embedding,  which is a tensor $\hbf_i^{(\ell)} \in \Rm^D$ resulting in a sequence of $b$-bit integers i.e., $\hbf_{i_{\tt INT}}^{(\ell)} \in {\{0,\dots,B-1\}}^D$. Instead of quantizing each node embedding, block-wise quantization takes a larger chunk of tensors and performs the quantization on them which further reduces the memory footprint and yields speedup. Block-wise quantization has been shown to be effective in reducing the memory footprint as demonstrated in~\cite{blockwise} where optimizer states are block-wise quantized to $8$-bits ({\tt INT8})\cite{8bitparallel}.

Consider the complete node embedding matrix after random projection, $\Hbf^{(\ell)}_\text{\tt proj} \in \Rm^{N\times R}$. To perform block-wise quantization first the node embedding matrix is reshaped into a stack of tensor blocks of length $G$:
\begin{equation}
    \Hbf_\text{\tt block}^{(\ell)} \in \Rm^{\frac{N\cdot R}{G} \times G} := \operatorname{reshape}\left(\Hbf^{(\ell)}_\text{\tt proj},G\right).
\end{equation}
The sequence of random projection and quantization as described in Section~\ref{sec:back} are performed on each block in $\hbf_{i_\text{\tt block}}^{(\ell)} \in \Rm^G \ \forall \ i = [1,\dots, (N\cdot R/G)]$. Performing quantization using larger blocks of tensors is shown to improve training stability, as block-wise quantization localizes the effects of outliers to within its own block~\cite{blockwise}. In this work, we experiment different block sizes to study the impact on memory consumption and test performance.

\vspace{-0.25cm}
\subsection{Improved Variance Minimization}
\label{sec:variance}
\begin{figure}[t]
    \centering
    \includegraphics[width=0.155\textwidth]{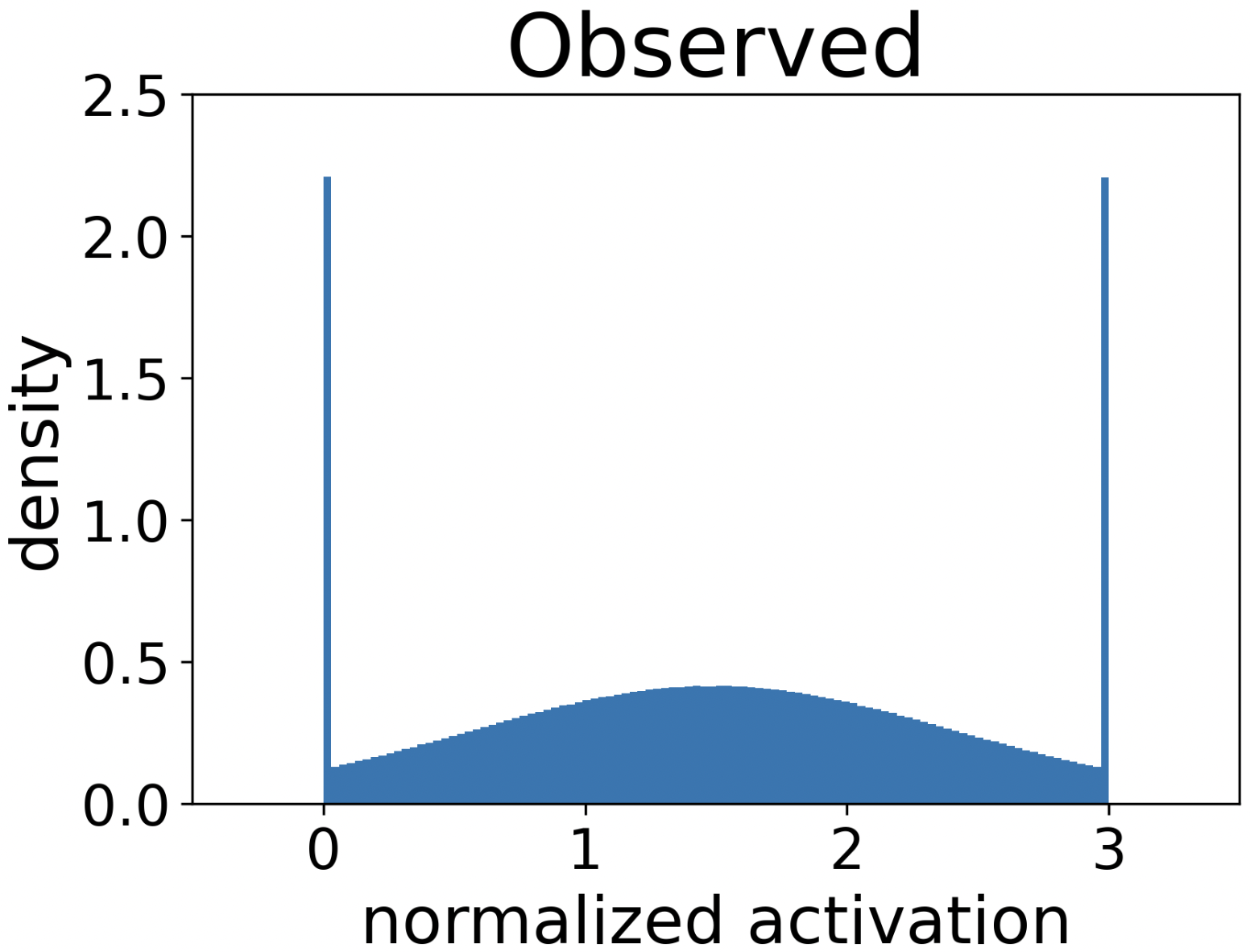}
    \includegraphics[width=0.155\textwidth]{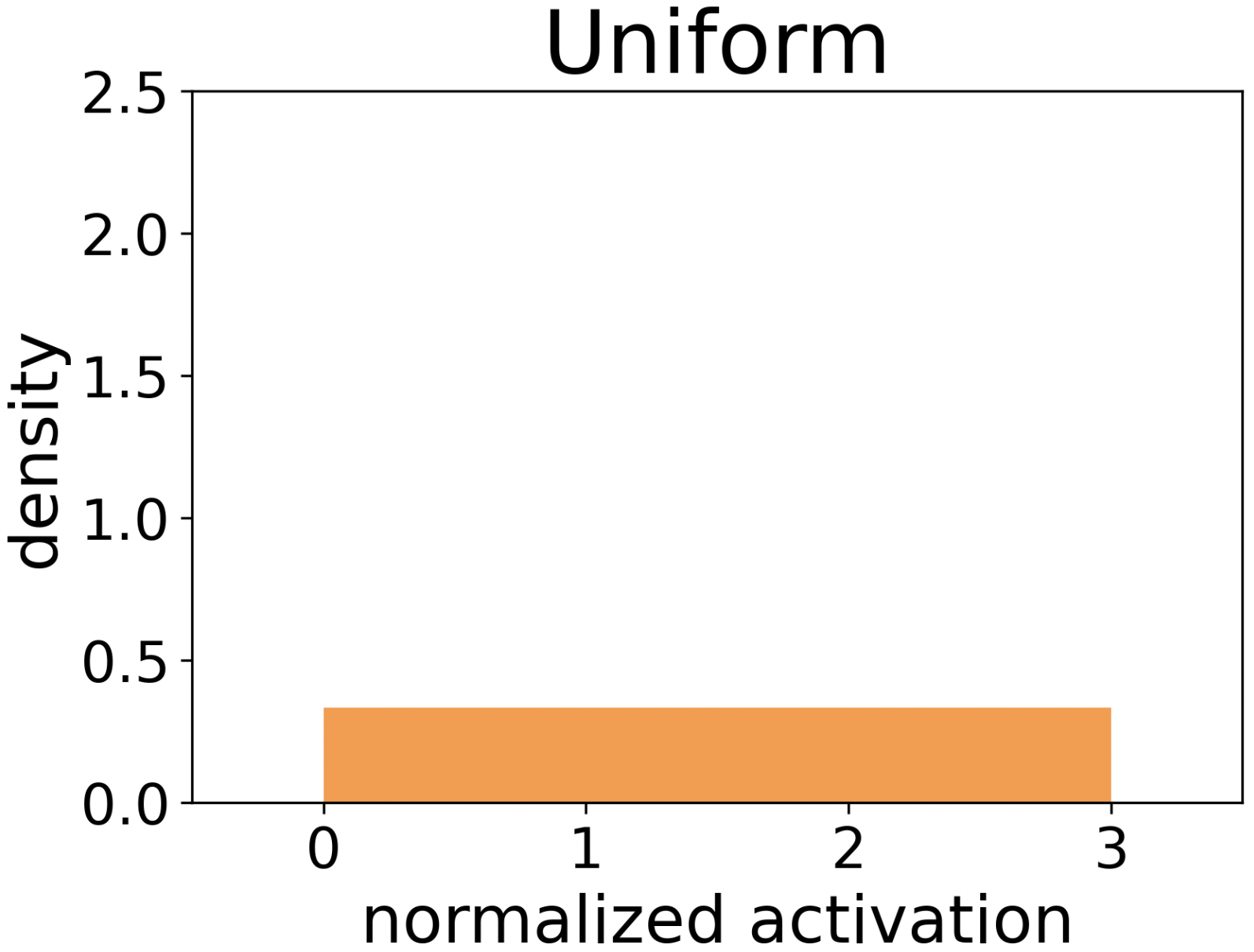}
    \includegraphics[width=0.155\textwidth]{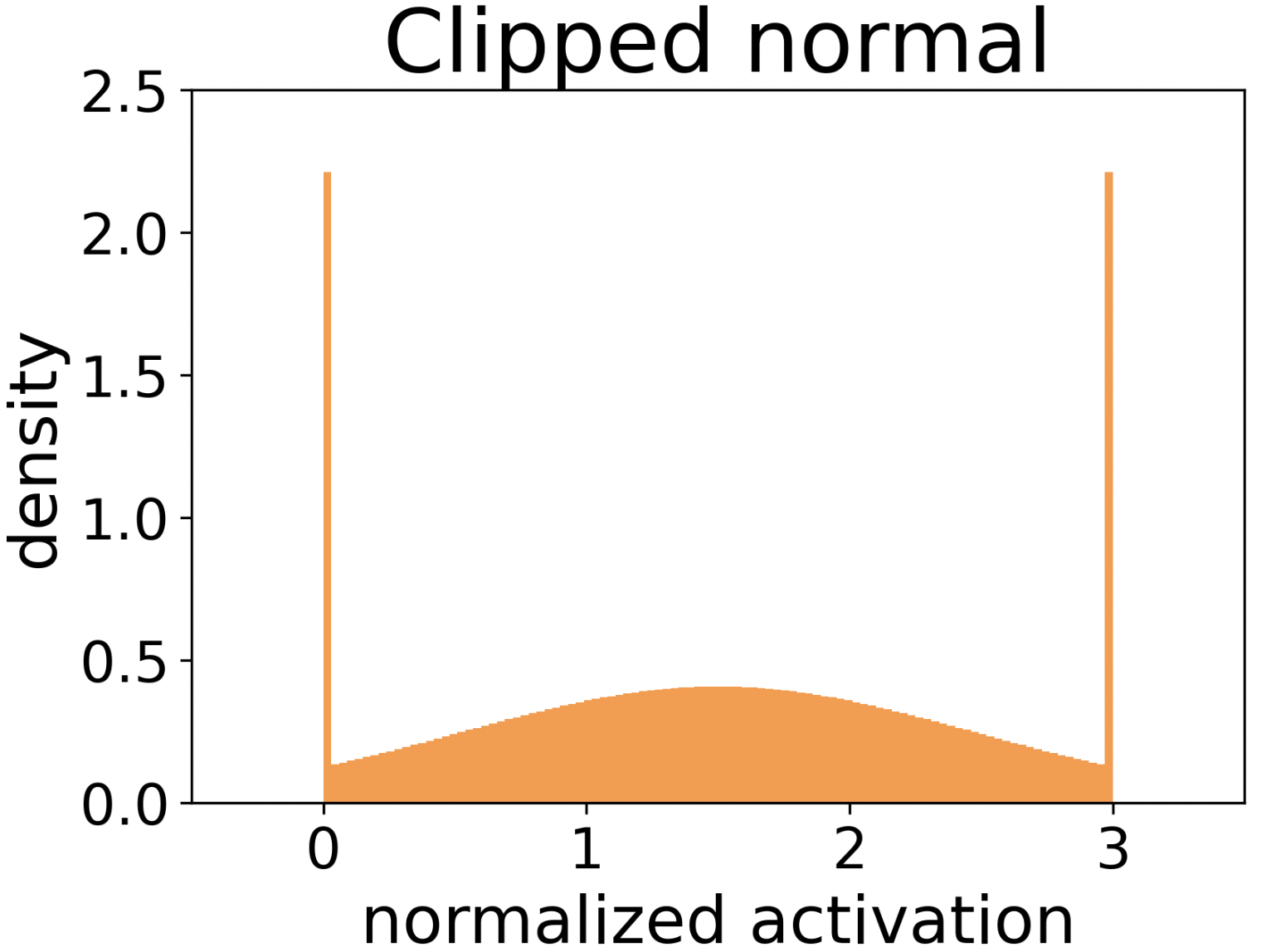}
    \vspace{-0.45cm}
    \caption{\textit{The observed normalized activations in a GNN model on the OGB-Arxiv data (left) compared to different modelled distributions: uniform (center), and clipped normal (right). Notice the clipped normal is able to model the observed distribution more accurately, including the edges where the spikes are caused due to clipping at the boundaries.}}
    \label{fig:all_dists}
    \vspace{-0.5cm}
\end{figure} 

Starting from the observation that $\hbf_{i_\text{\tt INT}}^{(\ell)}$ is an unbiased estimate, we want to find the quantization boundaries such that its variance, $\operatorname{Var}(\hbf_{i_\text{\tt INT}}^{(\ell)})$, is minimized to further reduce the effect of quantization. To achieve this we need three components: 1) distribution of activation maps, 2) variance as a function of the activation maps, and 3) minimization of the expected variance as a function of quantization boundaries. 


In the EXACT~\cite{exact}, the quantization boundaries are always set to integer values i.e., bins are of constant width. This stems from the assumption that the underlying distribution of activation maps are {\em uniformly} distributed~\cite{exact} (Figure~\ref{fig:all_dists}-center). In this work we show, on multiple datasets, that the activation maps are more accurately distributed as a variation of the normal distribution which we call the clipped normal. 

Letting $B=2^b-1$ define the number of quantization bins, and $\Phi^{-1}$ the Probability Point Function, we describe the clipped normal distribution as
\begin{align}
  \mathcal{CN}_{[1/D]}(\mu, \sigma) = \operatorname{min}\left(\operatorname{max}\left(0, \mathcal{N}(\mu, \sigma)\right), B\right), \label{eq:cn_dist}\\
  \text{where }\mu=B/2 \text{ and } \sigma = -\mu/\Phi^{-1}(1/D). \nonumber
\end{align}
The similarity between the observed and the modelled activation maps are visualized in Figure~\ref{fig:all_dists}, where we can see that the clipped normal distribution is better at approximating the activation maps compared to the uniform distribution.

We next expand SR to use irregular bin widths. Consider the normalized activation, $h \in \bar{\hbf}$ within the bin-$i$, stochastic rounding when using irregular bin widths, $\delta_i \ \forall \ i = [1,\dots,B]$, is given by:
\begin{equation}
    \left\lfloor h \right\rceil = \begin{cases}
        \lceil h \rceil, \text{with probability } (h-{\lfloor h \rfloor})/{\delta_i}\\
        \lfloor h \rfloor, \text{with probability } 1-((h-{\lfloor h \rfloor})/{\delta_i}).\\
    \end{cases}
    \label{eq:nonuinform_sr}
\end{equation}
Following the variance estimation from~\cite{improved_rounding}\footnote{Check Eq. 4.4 onwards in~\cite{improved_rounding} for detailed derivation.} and assuming a normalized activation $h$, we calculate its SR variance as 
\begin{equation}
\operatorname{Var}(\lfloor h \rceil) = \sum_{i=1}^{i=B} \left(\delta_i(h - \alpha_{i-1}) - (h - \alpha_{i-1})^2\right),  
\label{eq:sr_var}
\end{equation}
where $\delta_i$ is the width of the bin containing $h$, and $\alpha_i$ is the starting position of the bin.

Assuming {\tt INT2} quantization i.e., with $B=3$ bins, the expected variance of the SR operation under the clipped normal distribution is obtained from Eq.~\eqref{eq:sr_var} and Eq.~\eqref{eq:cn_dist}:
\begin{align}
&\mathbb{E}[\operatorname{Var}(\sr{h})] = \int_{0}^{\alpha} (\alpha\cdot h - h^{2}) \mathcal{CN}(h;\mu,\sigma) \, dh \nonumber \\ \nonumber
+& \int_{\alpha}^{\beta} \left((\beta - \alpha) (h - \alpha) - (h - \alpha)^{2}\right) \mathcal{CN}(h;\mu,\sigma) \, dh \\ 
+&\int_{\beta}^{B} \left((B - \beta) (h - \beta) - (h - \beta)^{2}\right) \mathcal{CN}(h;\mu,\sigma) \, dh
\label{eq:variance}
\end{align}
where $[\alpha,\beta]$ are the tunable edges of the central bin (see Figure~\ref{eq:quant}-B). Given this expected variance in Eq.~\eqref{eq:variance}, we optimize the boundaries $[\alpha,\beta]$ that minimize the variance due to SR. This can be done using standard numerical solvers implemented in Python.

\begin{figure}[t]
    \centering    \includegraphics[width=0.157\textwidth]{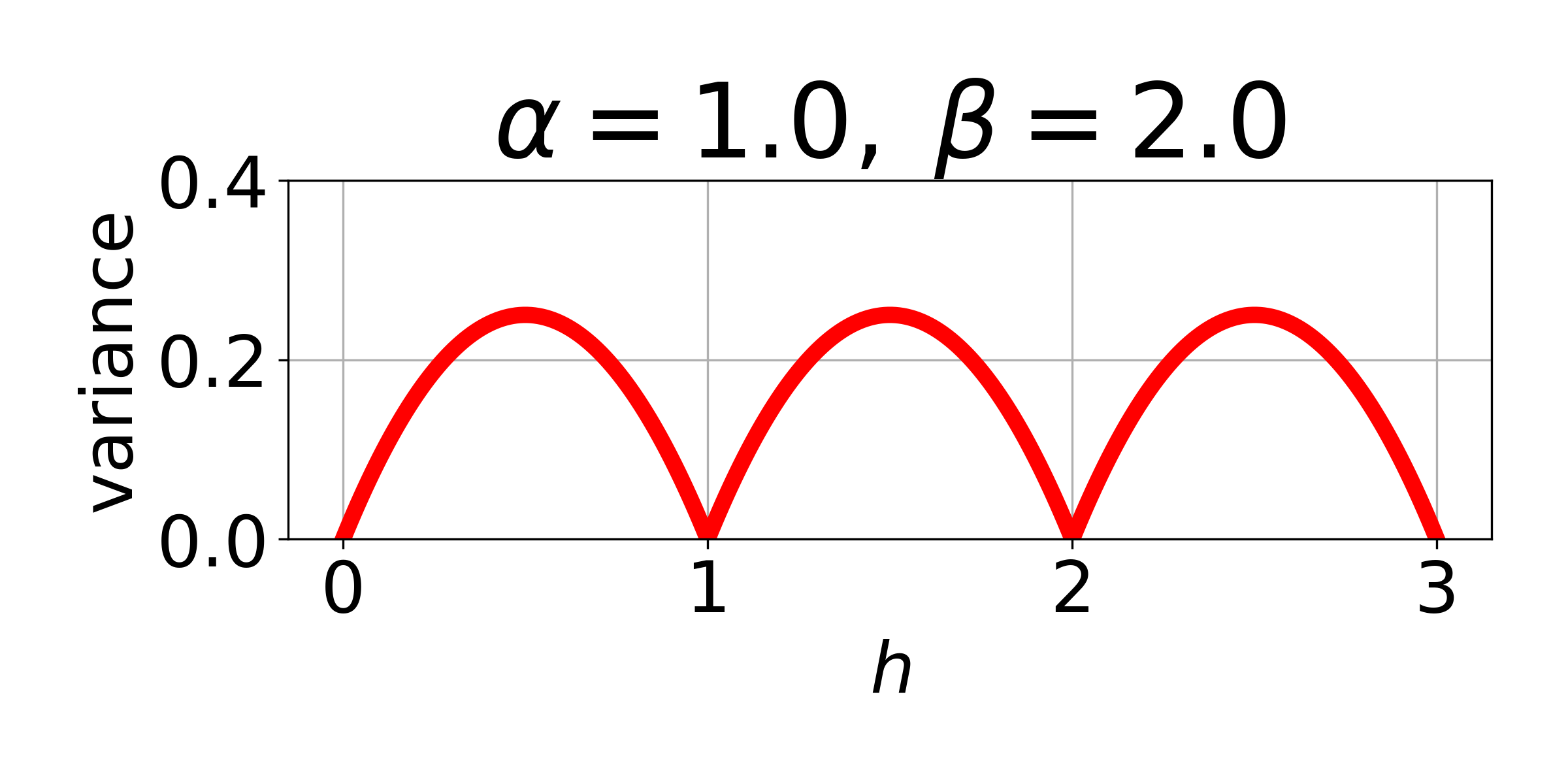}
    \includegraphics[width=0.157\textwidth]{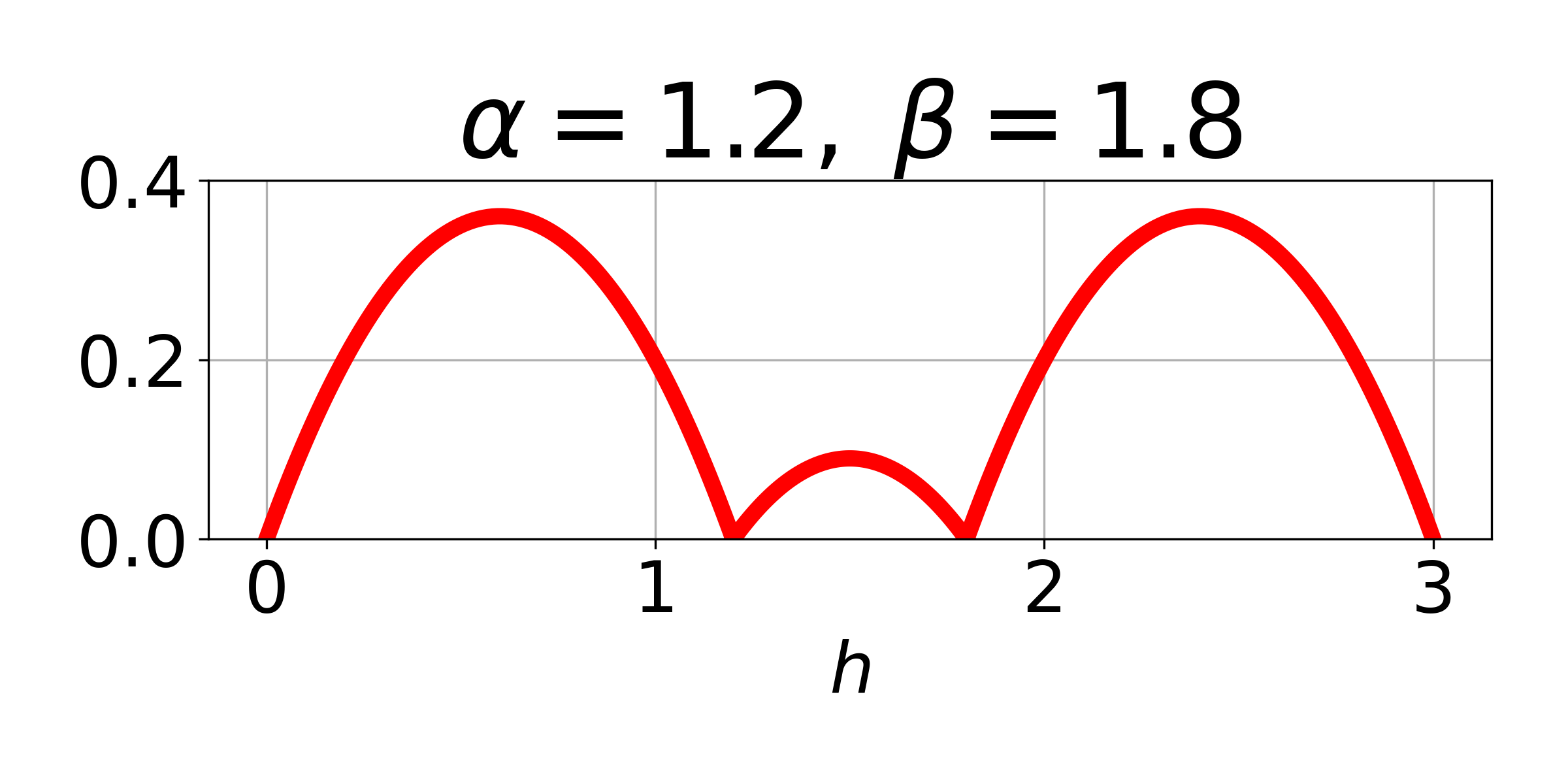}
    \includegraphics[width=0.157\textwidth]{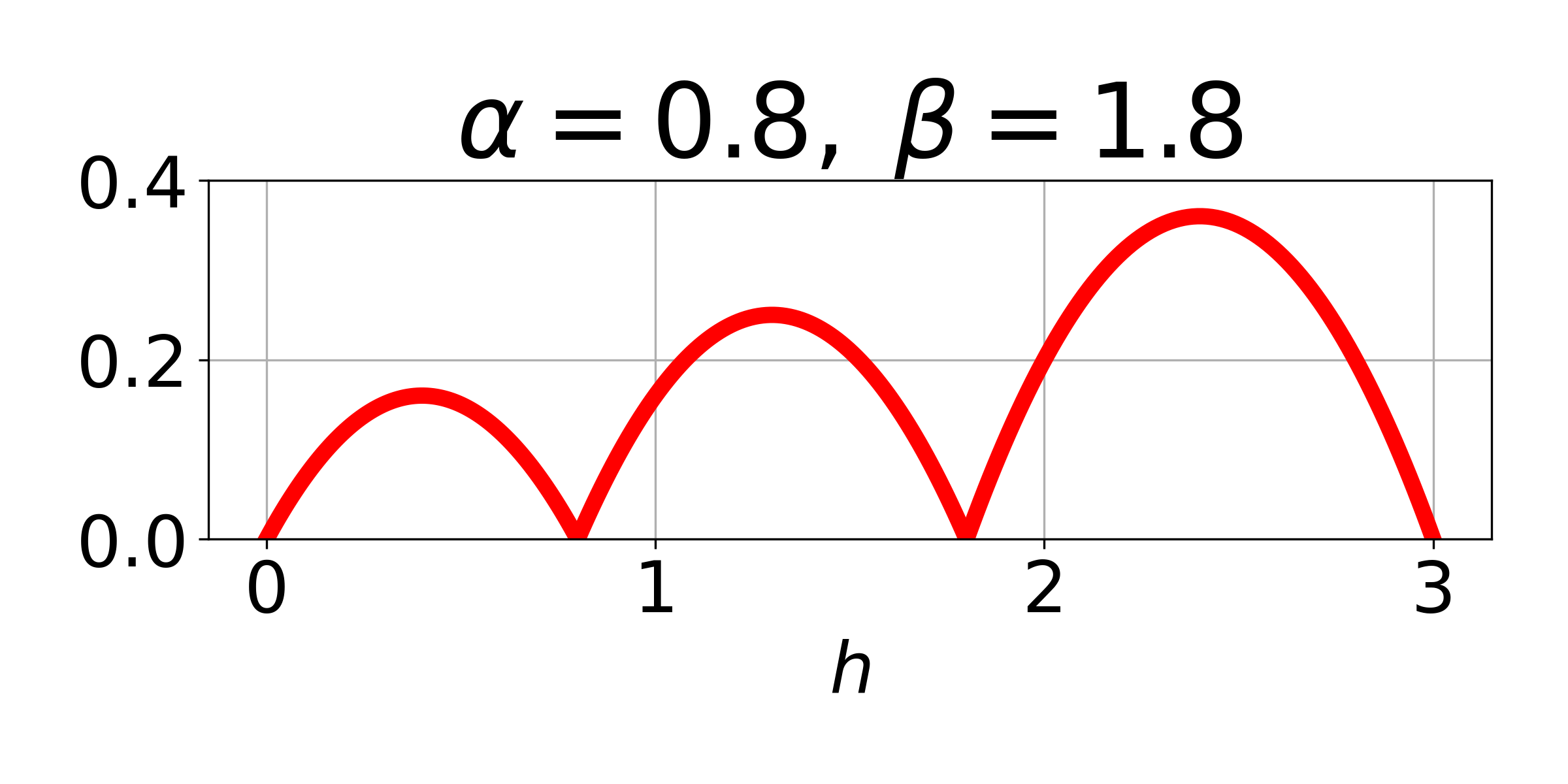}
    \vspace{-0.75cm}
    \caption{\textit{Variance of SR for {\tt INT2} quantization with different quantization boundaries $[\alpha,\beta]$ based on Eq.~\eqref{eq:sr_var}. When $[\alpha=1.0,\beta=2.0]$ uniform bin width is obtained.}}
    \label{fig:sr_vars}
    \vspace{-0.5cm}
\end{figure}

\vspace{-0.25cm}
\section{Experiments and Results}
\vspace{-0.25cm}
{\bf Data}: Experiments are performed on two large-scale graph benchmark datasets for inductive learning tasks. The open graph benchmark (OGB) Arxiv dataset~\cite{hu2020open} consisting of graph with $\approx 170k$ nodes and $>1M$ edges, and the Flickr dataset~\cite{fey2019pytorchgeometric} consisting of $\approx 90k$ nodes and $\approx 900k$ edges.\\
{\bf Experimental Set-up}: The GNN used in our experiments are the popular GraphSAGE architecture~\cite{hamilton2017inductive} implemented in
Pytorch~\cite{paszke2019pytorch}, which is also the baseline model with no activation compression i.e., operating in {\tt FP32} precision. EXACT is used in {\tt INT2} precision and $D/R = 8$ as the second baseline which uses extreme compression. We experiment our proposed compression methods in {\tt INT2} precision and different group sizes $G/R=[2,4,8,16,32,64]$ to demonstrate the influence of block-wise quantization. To keep the dimensionality proportion between the GNN layers, we scale the dimensionality of each layer equally when performing grouping, hence the block size is presented using the $G/R$. The influence of variance minimization (VM) on the test performance is also reported.  
\\
{\bf Results}: Performance of the baseline methods and different configurations of the method presented in this work for two datasets are reported in Table~\ref{tab:blockwise_int2}. The most astonishing trend is that there is no noticeable difference in test performance on both datasets, across all models, even with extreme quantization ({\tt INT2}) and any of the reported block sizes. With our proposed method there is a further improvement in memory consumption compared with EXACT by about 15\% (97\% with baseline {\tt FP32}) and about 8\% (97\% with baseline {\tt FP32}) for the Arxiv and Flickr datasets, respectively, when using the largest block size (G/R=64). We also gain a small speedup in training time per epoch: 5\% for Arxiv, and 2.5\% for Flickr, compared to EXACT. \\
\indent Use of clipped normal distribution in Eq.~\eqref{eq:cn_dist} to model the activation maps is better than uniform distribution. This is captured using the Jensen-Shannon divergence measure, reported in Table~\ref{table:cn_rv} where we observe that for all layers, in both datasets, the distance to the observed distribution is smaller for clipped normal distribution. \\
\indent Variance minimization does indeed decrease the variance induced by SR when performed with EXACT (Table~\ref{table:cn_rv}), but shows no change in performance (Table~\ref{tab:blockwise_int2}).

\vspace{-0.25cm}
\section{Discussion and Conclusion}
\vspace{-0.25cm}
Based on the experiments and the results in Table~\ref{tab:blockwise_int2}, we notice that block-wise quantization of activation maps on top of random projection and SR yields a further reduction in memory consumption and a small speedup in runtime. Increasing block size does not hamper the test performance but progressively yields further reduction in memory consumption.

\begin{table}[t]
\scriptsize
\centering
\begin{tabular}{ccccrr}
\toprule
{\bf Dataset} &{\bf Quant.} &{\bf G/R} & {\bf Accuracy} $\uparrow$ & {\bf S} (e/s) $\uparrow$ & {\bf M}(MB) $\downarrow$ \\ 
\midrule
\multirow{9}{*}{Arxiv} & {\tt FP32}~\cite{hamilton2017inductive} & -- & 71.95  $\pm$ 0.16 & 13.07            & 786.22         \\ 
& {\tt INT2}~\cite{exact} & -- & 71.16 ± 0.21 & 10.03 & 30.47 \\ 
 \cmidrule(l){2-6}

&\multirow{7}{*}{\tt INT2} & 2 & 71.16 ± 0.34 & 10.23 & 27.89 \\ 

&& 4 & 71.17 ± 0.22 & 10.46 & 26.60 \\
&& 8 & 71.21 ± 0.39 & 10.54 & 25.95 \\ 

& & 16 & 71.01 ± 0.19 & 10.55 & 25.72 \\ 

& & 32 & 70.87 ± 0.29 & 10.54 & 25.60 \\ 

& & 64 & 71.28 ± 0.25 & 10.54 & 25.56 
 \\ 
\cmidrule(l){2-6}
& {\tt INT2}+VM & -- & 71.20 ± 0.19 & 9.16 & 30.47 \\
\midrule \midrule
\multirow{9}{*}{Flickr} & {\tt FP32}\cite{hamilton2017inductive} & -- & 51.81  $\pm$ 0.16 & 17.95            & 546.92         \\ 
& {\tt INT2}\cite{exact} & -- & 51.65 ± 0.23 & 11.26 & 20.39 \\ 

 \cmidrule(l){2-6}

&\multirow{7}{*}{\tt INT2} & 2 & 51.58 ± 0.24 & 11.38 & 19.54 \\

& & 4 & 51.57 ± 0.29 & 11.50 & 19.12 \\ 

& & 8 & 51.60 ± 0.25 & 11.55 & 18.95 \\ 

& & 16 & 51.65 ± 0.21 & 11.54 & 18.86 \\ 

& & 32 & 51.61 ± 0.19 & 11.53 & 18.84 \\ 

& & 64 & 51.72 ± 0.24 & 11.53 & 18.84 \\ 
\cmidrule(l){2-6}
& {\tt INT2}+VM& -- & 51.71 ± 0.18 & 10.78 & 20.39 \\

\bottomrule
\end{tabular}
\vspace{-0.2cm}
\caption{\textit{Performance of block-wise quantization with $D/R = 8$, different quantization precision ({\tt FP32, INT2}), block size (G), and with variance minimization (VM). We report the following metrics on the {\em Arxiv}~\cite{hu2020open} and {\em Flickr}~\cite{fey2019pytorchgeometric} datasets: accuracy (\%), speed (S) reported as epochs/second and memory (M) consumption in MB. Standard deviations of test accuracy is computed over 10 runs.}}
\label{tab:blockwise_int2}
\vspace{-0.15cm}
\end{table}

\begin{table}[t]
\centering
\scriptsize
\begin{tabular}{lcccccc}
\toprule
\textbf{Dataset} & \textbf{Layer} & {\bf R} &  $\mathcal{U}$ & $\mathcal{CN}_{[1/D]}$ & \textbf{Var. Reduction} $(\%)$ \\ \midrule
\multirow{2}{*}{Arxiv} 
& layer 1 &  16 & 0.0495 & { 0.0213} & 3.17\\
& layer 2 &  16 & 0.0446 & { 0.0016} & 2.09 \\
& layer 3 &  16 & 0.0451 & { 0.0041} & 2.19\\ 
\midrule
\multirow{2}{*}{Flickr} 
& layer 1 &  63 & 0.0674 & {0.0017} & 6.14\\
& layer 2 &  32 & 0.0504 & {0.0033} & 4.37\\ 
\bottomrule
\end{tabular}
\vspace{-0.2cm}
\caption{\textit{Jensen-Shannon divergence measure for Uniform and Clipped Normal distributions compared to the normalized activations $\bar{\hbf}$ at each layer of the GNN for Arxiv and Flickr datasets. In all cases we see a smaller divergence measure between the clipped normal and the empiricial distribution of activation maps.}}
\label{table:cn_rv}
\vspace{-0.5cm}
\end{table}

Activation maps in GNNs are not uniformly distributed; we demonstrated this using empirical visualizations in Figure~\ref{fig:all_dists}. We quantified this using the clipped normal distribution which had a smaller Jensen-Shannon divergence to the observed distribution, as seen in Table~\ref{table:cn_rv}. This implies that using uniform quantization bin width could be sub-optimal. We presented an extension to stochastic rounding that accounts for variable bin widths in Eq.~\eqref{eq:nonuinform_sr}. The influence on quantization variance using Eq.~\eqref{eq:sr_var} visualized in Figure~\ref{fig:sr_vars} clearly demonstrates the value of using non-uniform bin widths.
\\
{\bf Limitations}: The compute overhead even with the proposed modifications do not fully recover the reduction in speedup compared to the baseline i.e., when using {\tt FP32}. While the variance estimation improvement introduced by modelling the activation maps with clipped normal distribution better models the activation maps, minimizing the variance of SR under this distribution does not yield a noticeable improvement in test performance. This could simply be due to the fact that the overall drop in performance even with block-wise quantization is small, and there is no room for further improvement. The software implementations of the quantization and variance minimization strategies are not highly optimized and there is room for further fine-tuning.
\\
{\bf Conclusion}: Improving efficiency of training GNNs is an important step towards reducing their resource consumption. We have demonstrated that combining block-wise quantization with extreme compression (down to {\tt INT2}) can be achieved with a small drop performance. The reduction in memory consumption from baseline ({\tt FP32}) is $>95$\%, and compared to EXACT we gain a further $>15$\% in memory reduction and up to $\approx 5$\% training runtime speedup per epoch. We have empirically shown that the activation maps for common GNN architectures do not follow uniform distribution. We proposed an improved modelling of these activation maps using a variation of the normal distribution (the clipped normal) and show that tighter variance minimization of the quantization noise was achievable.

\balance
\bibliographystyle{IEEEbib}
\bibliography{strings,refs}

\input{appendix}
\end{document}

%% file: Macros.tex
\def \Rm{{\mathbb{R}}}
\def \Em{{\mathbb{E}}}

\def \Abf{{\mathbf A}}

\def \Dbf{{\mathbf D}}

\def \hbf{{\mathbf h}}
\def \Hbf{{\mathbf H}}

\def \Ibf{{\mathbf I}}

\def \Rbf{{\mathbf R}}

\def \Xbf{{\mathbf X}}

\def \Hbf{{\mathbf H}}
\def \0bf{{\mathbf 0}}
\def \Hbfp{{\overline{\Hbf}^{(\ell)}_\text{\tt proj}}}

\def \Gcal{{\mathcal G}}

\newcommand{\sr}[1]{\lfloor #1 \rceil}

%% file: appendix.tex
\appendix
\section{Stochastic Rounding with non-uniform bins}
In this section, we elaborate on stochastic rounding when using non-uniform bin widths. Consider the normalized activation, $h \in \bar{\hbf}$ within the bin-$i$, stochastic rounding when using irregular bin widths, $\delta_i \ \forall \ i = [1,\dots,B]$, is given by:
\begin{equation}
    \left\lfloor h \right\rceil = \begin{cases}
        \lceil h \rceil, \text{with probability } (h-{\lfloor h \rfloor})/{\delta_i}\\
        \lfloor h \rfloor, \text{with probability } 1-\left((h-{\lfloor h \rfloor})/{\delta_i}\right).\\
    \end{cases}
    \label{eq:nonuinform_sr_app}
\end{equation}
The key difference with stochastic rounding with uniform width is that the probability now has a scaling of the bin width, $\delta_i$, to account for the different bin widths. 

We first show that stochastic rounding with non-uniform width is unbiased. Consider, the expectation according to~\eqref{eq:nonuinform_sr_app}:
\begin{align}
    \Em[\left\lfloor h \right\rceil] &= \lceil h \rceil \cdot (h-{\lfloor h \rfloor})/{\delta_i} + \lfloor h \rfloor \cdot (1-\left((h-{\lfloor h \rfloor})/{\delta_i}\right)) \nonumber \\ 
    &\text{Rearranging the terms, } \nonumber \\
    &= (\lceil h \rceil - \lfloor h \rfloor) (h-{\lfloor h \rfloor})/{\delta_i} + \lfloor h \rfloor \nonumber \\
    &\text{Observing that } \delta_i=(\lceil h \rceil - \lfloor h \rfloor)
    \nonumber \\
    &= (h-{\lfloor h \rfloor}) + \lfloor h \rfloor \nonumber \\
    \Em[\left\lfloor h \right\rceil] &= h, \nonumber
\end{align}
which proves that stochastic rounding with non-uniform bin widths is also an unbiased estimate. 

In order to estimate the variance of stochastic rounding with uniform bin width, $\delta$, we follow the variance estimation from~\cite{improved_rounding} with an unknown probability $p$ of rounding :
\begin{equation}
\operatorname{Var}(\lfloor h \rceil) = \delta^2\left(p - p^2\right).
\label{eq:sr_var_improv}
\end{equation}
When considering non-uniform bin widths, the variance estimation in Eq.~\eqref{eq:sr_var_improv} should be modified to account for the difference in bin width and the corresponding change in the rounding probabilities, resulting in:
\begin{equation}
\operatorname{Var}(\lfloor h \rceil) = \sum_{i=1}^{i=B} \delta_i^2\left((h - \alpha_{i-1})/\delta_i - \left((h - \alpha_{i-1})/\delta_i\right)^2\right),  
\label{eq:sr_var_app}
\end{equation}
where we have used the probabilities scaled by the bin width from Eq.~\eqref{eq:nonuinform_sr_app}, $\delta_i$ is the width of the bin-$i$. Eq.~\eqref{eq:sr_var_app} can be simplified to:
\begin{equation}
\operatorname{Var}(\lfloor h \rceil) = \sum_{i=1}^{i=B} \left( \delta_i(h - \alpha_{i-1}) - \left((h - \alpha_{i-1})\right)^2\right)
\label{eq:sr_var_app}
\end{equation}
reported in Eq.~\eqref{eq:sr_var}.

Assuming {\tt INT2} quantization i.e., with $B=3$ bins and adjustable central bin, we have $[\alpha_0 = 0, \alpha_1=\alpha,\alpha_2=\beta, \alpha_3=3]$. 
We next work out the variance for each of the three bins by substituting the bin-specific values in Eq.~\eqref{eq:sr_var_app}: 
For $\delta_1=\alpha$ when $ 0 \leq h < \alpha$ with :
\begin{equation}
\operatorname{Var}(\lfloor h \rceil;\delta_1) = (\alpha \cdot h - h^{2})  
\end{equation}
For $\delta_2=\beta-\alpha$ when $ \alpha \leq h < \beta$ with :
\begin{equation}
\operatorname{Var}(\lfloor h \rceil;\delta_1) = (\beta-\alpha) (h - \alpha) - (h-\alpha)^{2} 
\end{equation}
For $\delta_3=B-\beta$ when $ \beta \leq h \leq B$ with :
\begin{equation}
\operatorname{Var}(\lfloor h \rceil;\delta_1) = (B-\beta) (h - \beta) - (h-\beta)^{2} 
\end{equation}

Using these variance estimates for each bin, the expected variance of the stochastic rounding operation under the clipped normal distribution is obtained Eq.~\eqref{eq:cn_dist}:
\begin{align}
&\mathbb{E}[\operatorname{Var}(\sr{h})] = \int_{0}^{\alpha} (\alpha \cdot h - h^{2}) \mathcal{CN}(h;\mu,\sigma) \, dh \nonumber \\ \nonumber
+& \int_{\alpha}^{\beta} \left((\beta - \alpha) (h - \alpha) - (h - \alpha)^{2}\right) \mathcal{CN}(h;\mu,\sigma) \, dh \\ 
+&\int_{\beta}^{B} \left((B - \beta) (h - \beta) - (h - \beta)^{2}\right) \mathcal{CN}(h;\mu,\sigma) \  dh.
\label{eq:variance_app}
\end{align}


\vspace{-0.35cm}
\section{Tuning quantization boundaries in practice}
\label{app:quant_tuning}


Currently the bin boundaries for \texttt{INT2} quantization  ($\alpha$ and $\beta$) are obtained by the minimization of Eq.~\eqref{eq:variance} numerically. This is because of the complex nature of the integral. Since we know that for $D>2048$ we get out-of-memory errors, we only need to calculate the optimal boundaries of Eq.~\eqref{eq:variance} for the cases where the distribution is $\mathcal{CN}_{[1/D]}$ with $D\in\{4,5...,2048\}$.\footnote{We calculate all these cases in \url{https://github.com/saintslab/i-Exact/blob/main/var_opt.ipynb}.} 
This now gives us the ability to map directly from the dimensionality of the projected layers {R} to the optimal boundaries, according to variance minimization. 

\vspace{-0.35cm}
\section{Validation of Variance Minimization}
To test if variance minimization also works in practice we look at the normalized activations after projection $\Hbfp$ and attempt to characterise the variance induced by SR. If we denote $\alpha^*$ and $\beta^*$ as the quantization boundaries that minimize Eq.~\eqref{eq:variance} and we denote uniform SR as $\lfloor \cdot \rceil$, and SR with the optimized boundaries $\alpha^*$ and $\beta^*$ as
$\lfloor \cdot \rceil^*$. We can calculate the reduction in variance as follows:
\begin{equation}
\text{Var. Reduction} = 
1 - \frac{\sum\left[\left(\Hbfp - \Big\lfloor \Hbfp \Big\rceil^*\right)^2\right]}{\sum\left[\left({\Hbfp - \Big\lfloor \Hbfp} \Big\rceil\right)^2\right]}.
\end{equation}

\vspace{-0.25cm}
\begin{figure}[h]
    \centering
    \includegraphics[width=0.35\textwidth]{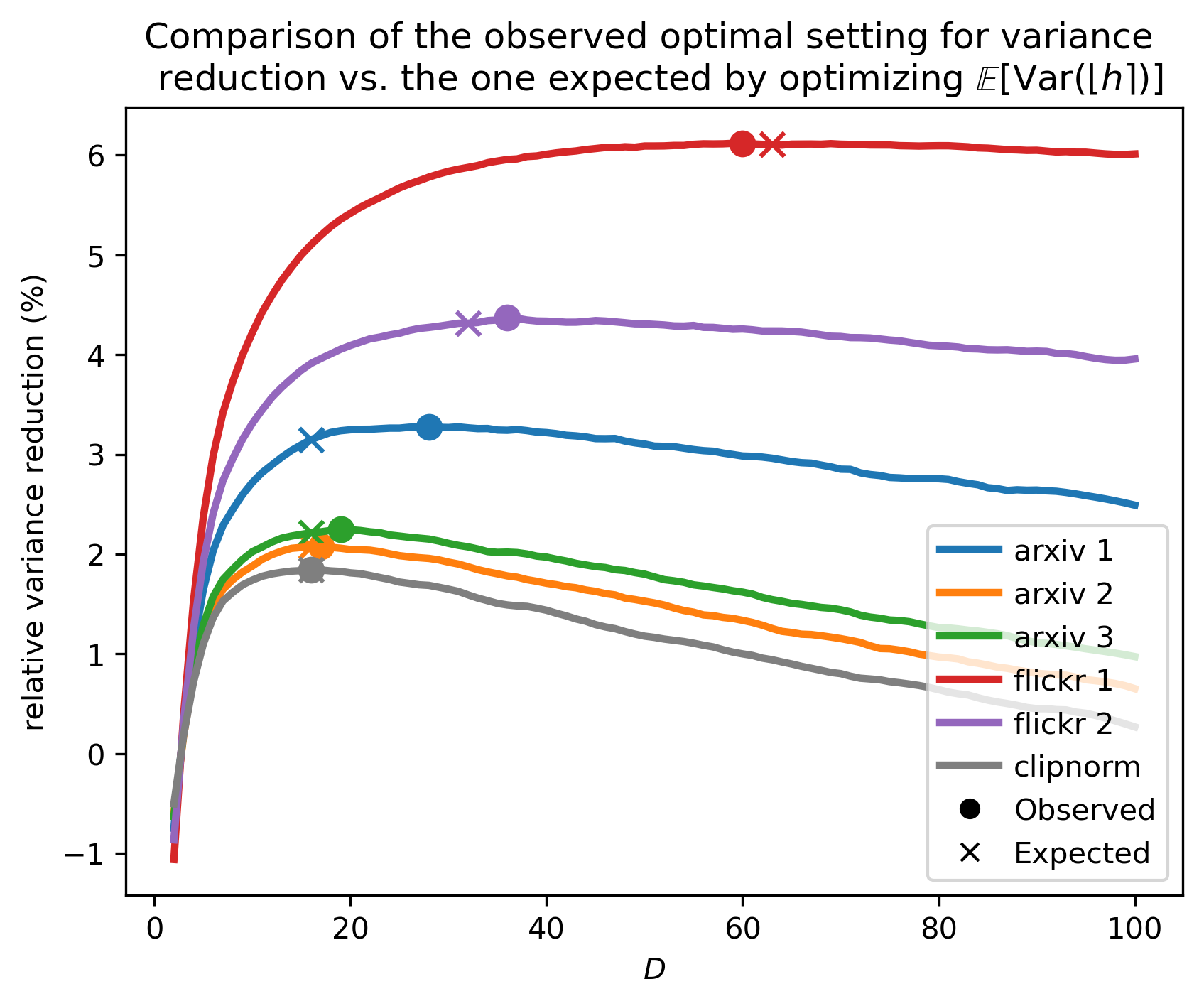}
    \vspace{-0.35cm}
    \caption{This plot demonstrates the relative variance reduction across different layers $\Hbfp$ and a clipnormal distribution model when performing variance minimization. Here the crosses indicate what we would expect the optimal dimensionality parameter for variance minimization to be, and the circles indicate what it actually is.}
    \vspace{-0.25cm}
    \label{fig:layer_var_reductions}
\end{figure}

To see whether the quantization boundaries $\alpha^*$ and $\beta^*$ are actually the boundaries that result in the optimal variance reduction, we compare the observed maximal variance reduction with the one we would expect from the dimensionality of $\Hbfp$, also known as $R$. $\alpha^*$, and symmetrically $\beta^*$, are found and used as follows:
\begin{enumerate}
    \item Given a distribution of activations $\Hbfp$, we naively find the clipnorm distribution that best approximates it. This is always $\mathcal{CN}_{[1/{R}]}$, since this ensures that the density of the outer quantization boundaries $0$ and $3$, are equal for the actual- and theoretical distributions.
    \item In accordance with the method in Appendix~\ref{app:quant_tuning}, we find the inner quantization boundaries that minimize the variance of Eq.~\eqref{eq:variance}, with the dimensionality parameter set to $R$ which is referred to as $\alpha^*$ and $\beta^*$.
    \item We then set these as our new quantization boundaries, instead of the integer quantization boundaries, used in uniform quantization.
\end{enumerate}

The variance reduction achieved by the aforementioned method is what we refer to as the {\em maximal expected variance reduction}, since our naive assumption is that we \textit{expect} all activations to be distributed as a clipnorm. Our maximal observed variance reduction corresponds to when using a clipnorm the dimensionality obtained from Appendix~\ref{app:quant_tuning} as the parameter. The reason that these two numbers can differ, is both the fact that our assumption that activations are distributed under the clipped normal is not always true, and also that there is an inherent stochasticity when using SR. This is captured in Figure~\ref{fig:layer_var_reductions}, where we see that variance minimization is maximal when we assume the dimensionality to be $D$ that the clipnorm was defined from; in this case we have $D=16$ and thus the distribution $\mathcal{CN}_{[1/16]}$.

On the other hand there are some layers, like \texttt{arxiv 1} in which the expected and observed optimal setting for variance minimization are not very close. This might be due to the fact that the distribution of activations in \texttt{arxiv 1} is not that closely approximated by a clipped normal, which is also shown in Table~\ref{table:cn_rv}.

To further test if variance minimization always finds the optimal quantization boundaries, we can look at different clipnorm distributions. Specifically, the five $\mathcal{CN}_{[1/D]}$ distributions, with $D\in\{16, 32, 64, 96, 128\}$. 

\begin{figure}[h]
    \centering
    \includegraphics[width=0.35\textwidth]{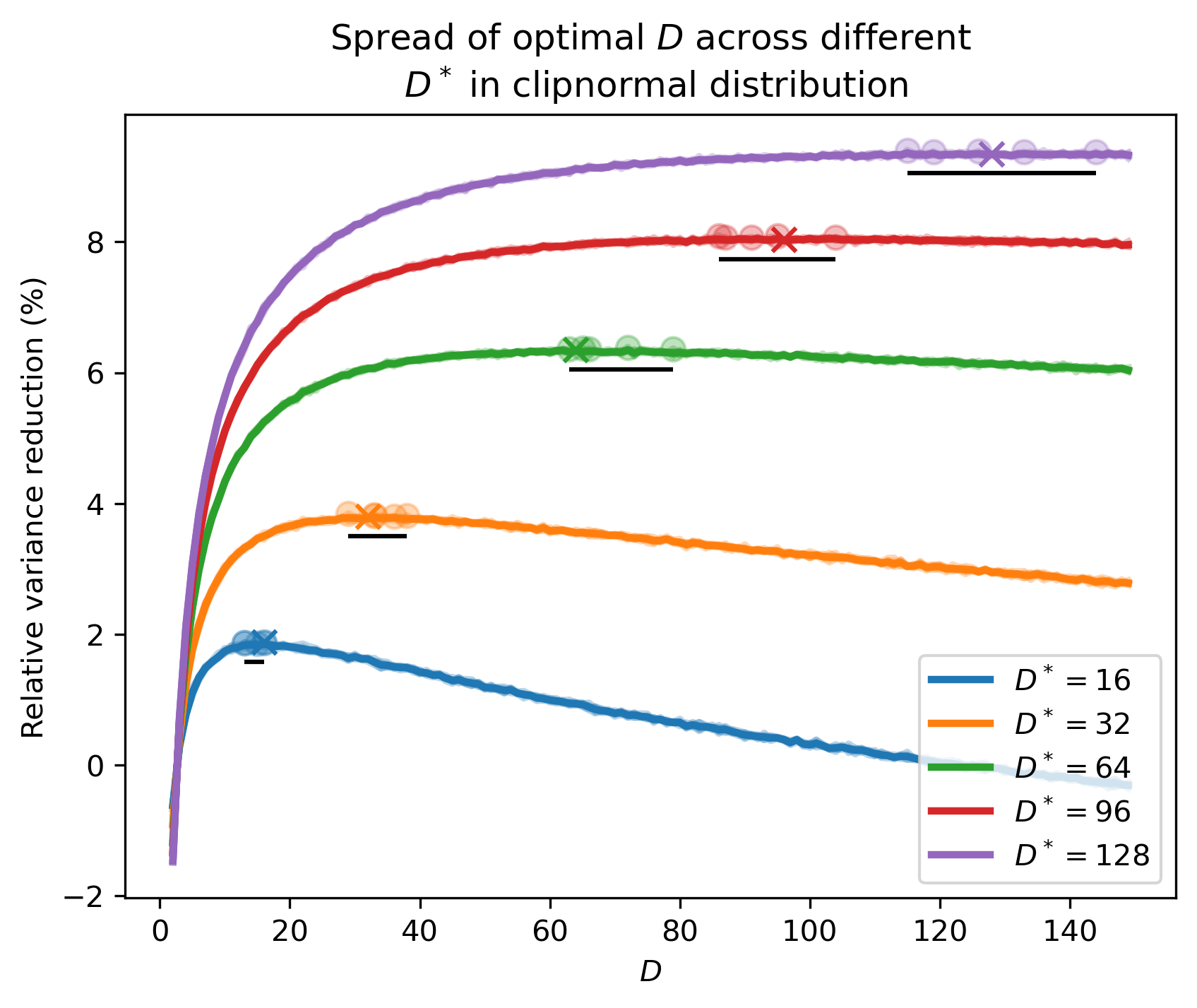}
    \vspace{-0.5cm}
    \caption{This plot illustrates the relative reduction in variance achieved through optimizing the assumed dimensionality in a clipnorm distribution, as indicated by different '$D\#$' hyperparameters. Each line represents the mean variance reduction across multiple trials for a given dimension, with shaded areas showing the range from minimum to maximum variance reduction observed. The markers on the line show the observed and expected maxima, highlighting the performance and consistency of the optimization across dimensions. The horizontal solid lines indicate the spread of observed maxima, providing insights into the variability and stability of the optimization process for each hyperparameter.}
    \vspace{-0.25cm}
    \label{fig:clipnorm_var_reductions}
\end{figure}

From Figure~\ref{fig:clipnorm_var_reductions} we can see that our observed optimal $D$ are always around the expected optimal $D$. We also see that the spread usually increases as the dimensionality increases. This might be because of the fact that the curves {\em level out} which means that only minor deviations, can change the actual optimal $D$ for variance minimization. The fact that our calculated or observed optimal $D$ for variance minimization are around the expected $D$ support the correctness of Eq.\eqref{eq:variance} and thus of variance minimization.

\vspace{-0.25cm}

\vspace{-0.25cm}
\section{Experimental Details}
\vspace{-0.25cm}
The experiments can be split into two main parts, the experiments in Table \ref{tab:blockwise_int2} and those in Table \ref{table:cn_rv}. To conduct the experiments in Table \ref{table:cn_rv}, we save the activation maps before quantization but after random projection, using the original EXACT configurations\footnote{Obtained from \url{https://github.com/zirui-ray-liu/Exact}}. After training using these configurations, we take the test activations belonging to the epoch with the lowest validations loss. We repeat this 10 times and take the mean and use this to define the {\em observed density} of activations as seen in Figure \ref{fig:all_dists}. The code for this paper is built as an extension of the EXACT repository and can be accessed on the link \url{https://github.com/saintslab/i-Exact}.